\documentclass{article}
\usepackage{spconf,amsmath,graphicx}
\usepackage{amsfonts}
\usepackage{graphicx}
\usepackage{xcolor}
\usepackage{times}
\usepackage{algorithm}
\usepackage{algpseudocode}

\title{Representation Decomposition for Image Manipulation and Beyond}
\name{Shang-Fu Chen$^{\star}$, Jai-Wei Yan$^{\star}$, Ya-Fan Su$^{\dag}$, Yu-Chiang Frank Wang$^{\star}$}
\address{$^{\star}$ Graduate Institute of Communication Engineering, National Taiwan University, Taiwan\\
$^{\dagger}$ Telecommunication Laboratories, Chunghwa Telecom Co., Ltd., Taiwan}

\begin{document}
%

\maketitle
\begin{abstract}
Representation disentanglement aims at learning interpretable features, so that the output can be recovered or manipulated accordingly. While existing works like infoGAN~\cite{chen2016infogan} and AC-GAN~\cite{odena2017icml} exist, they choose to derive disjoint attribute code for feature disentanglement, which is not applicable for existing/trained generative models. In this paper, we propose a decomposition-GAN (dec-GAN), which is able to achieve the decomposition of an existing latent representation into content and attribute features. Guided by the classifier pre-trained on the attributes of interest, our dec-GAN decomposes the attributes of interest from the latent representation, while data recovery and feature consistency objectives enforce the learning of our proposed method. Our experiments on multiple image datasets confirm the effectiveness and robustness of our dec-GAN over recent representation disentanglement models.
\end{abstract}
\section{Introduction}
Recent developments of Generative Adversarial Network (GAN)~\cite{goodfellow2014generative} models result in promising progresses and achievements in image generation.
In order to produce image outputs with desirable attributes (e.g., gender, expression, etc.), feature disentanglement aims at decomposing the above latent representation into distinct parts, each corresponding to particular properties. 
Representation disentanglement~\cite{chen2016infogan, odena2017icml, higgins2017iclr, kingma2014nips, kulkarni2015nips, kulkarni2015deep, liu2018cdrd,  sohn2015nips, zheng2019disentangling, press2020emerging, liu2018ufdn} aims at learning an interpretable representation from image variants, which can be realized in unsupervised or supervised settings. For example, with supervision of labeled data, AC-GAN~\cite{odena2017icml} factorizes representations into disjoint parts describing visual content and attribute information, respectively (e.g., image~\cite{isola2017pix2pix}, text~\cite{reed2016GAT,zhang2017stackgan}) during training. 
On the other hand, if training data are unlabeled, infoGAN~\cite{chen2016infogan} performs representations disentanglement by maximizing the mutual information between latent variables and data variation.

Despite promising performances, these works are not able to be directly applied on \textit{existing/pre-trained} generative models. In other words, their disentanglement mechanisms must be determined and trained in advance. More specifically, their need of deciding image attributes to be disentangled beforehand makes their feature disentanglement less flexible. If the attributes of interest are changed, the above generative models need to be trained from scratch again. Moreover, with the scale of generative models growing, training of state-of-the-art generative models become very time and resource-consuming. 




Instead of explicitly decomposing latent representation into disjoint parts, we propose a unique decomposition-GAN (dec-GAN) for performing feature disentanglement. Our disentanglement mechanism focuses on extracting attributes of interest (e.g., pose, expression, etc.) from latent representation, while the generator is \textit{fixed}. Depending on the attribute of interest, dec-GAN is guided by an attribute classifier trained for distinguishing the attribute. Together with image recovery objectives, dec-GAN decomposes visual features from a joint latent representation into separate ones associated with content and attribute of interest.
While recent works like~\cite{voynov2020unsupervised} and~\cite{shen2020interfacegan} deal with the similar task that learns disentangled features based on existing generative models, both of their methods are not able to manipulate particular attributes of interest when taking images as input.
On the other hand, with the above disentangled features, our dec-GAN is able to utilize existing generative models for describing each type of disentangled features, which allows improved and interpretable feature representations for image manipulation, along with additional flexibility in determining the attributes of interest after the generator is trained.

We now highlight the contributions of work as follows: 
\begin{itemize}
    \item We propose a novel learning scheme for representation disentanglement, which uniquely decomposes features of existing GAN-based models into interpretable representations. 
    \vspace{-0.15cm}
    \item Our learning framework does \textit{not} require pre-determined or disjoint latent representations to describe attributes in advance, and thus exhibits additional flexibility in determining the attributes of interest.
    \vspace{-0.15cm}
    \item Our experiments confirm that our model successfully decomposes latent features derived by existing GANs for image manipulation and classification.
\end{itemize}
\begin{figure*}
\begin{center}
\includegraphics[width=0.85\textwidth]{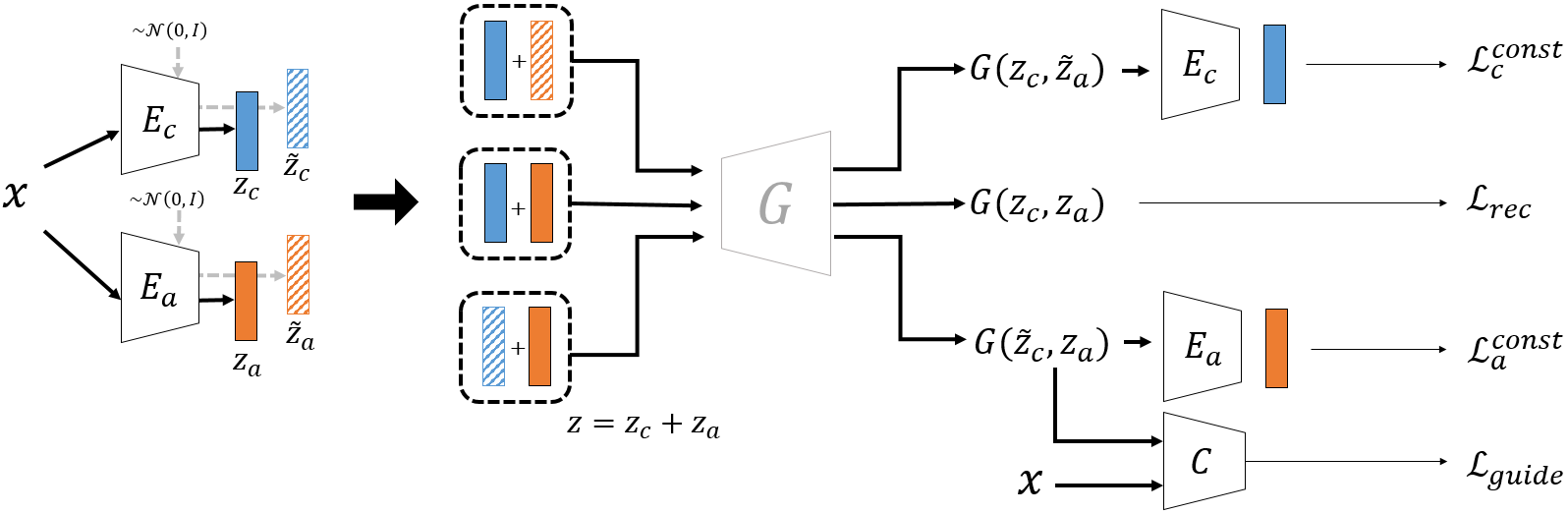}
\end{center}
\vspace{-0.5cm}
\caption{
Overview and architecture of our Decomposition-GAN (dec-GAN), which consists of content encoder $E_c$, attribute encoder $E_a$, and an auxiliary guidance attribute classifier $C$, while generator $G$ is fixed. Note that our dec-GAN decomposes latent features $z$ into separate representations (instead of disjoint ones), i.e., $z = z_c + z_a$. Note that $G(z_{c}, {\tilde{z}}_{a})$, $G(z_c, z_a)$ and $G({\tilde{z}}_c, z_a)$ indicate the image outputs synthesized from pairs of the associated content and attribute features.}
\label{fig:archi}
\end{figure*}
\section{Decomposition-GAN (dec-GAN) for Disentanglement}

We propose \textit{decomposition-GAN (dec-GAN)} for representation disentanglement. As illustrated in Figure~\ref{fig:archi}, our dec-GAN decomposes the latent code $z$ into content code $z_c$ and attribute code $z_a$ while satisfying $z = z_c + z_a$. In other words, based on existing latent feature $z$, our goal is to decompose it into content and attribute representations $z_c$ and $z_a$. We utilize two separate encoders $E_c$ and $E_a$ for extracting $z_c$ and $z_a$, respectively. The reconstruction output is denoted as $G(z_c, z_a) = G(z | z = z_c + z_a)$. It is worth noting that, as verified in Section~\ref{exp}, our dec-GAN can utilize existing state-of-the-art generative models.
\subsection{Attribute Guidance for Disentanglement}

In our dec-GAN, we first utilize the idea of data recovery to encourage generated images to be sufficiently realistic. For this reconstruction loss, we consider the L1 distance between the reconstructed and input images: 
\begin{equation}
\label{eq_rec}
L_{rec} = |G(z_c, z_a)-x|.
\end{equation}

Following VAE-GAN~\cite{larsen2015autoencoding} and DRIT~\cite{lee2018drit}, we fit the distributions of encoded content and attribute features to normal distributions, which allow improved/continuous data representation ability. This can be achieved by minimizing the Kullback–Leibler divergence (KLD) between each distribution and $\mathcal{N}(0,1)$. However, since the disentangled content and attribute features describe distinct information, we do not expect them to fit the same normal distribution. Therefore, we calculate the KLD loss for each feature as follows, 
\begin{equation}
\label{eq_kl_c}
L_{KL,c} = \mathbb{E}[KL(P(z_c')||\mathcal{N}(0,1))], z_c = E^{fc}_c(z_c'),
\end{equation}
\begin{equation}
\label{eq_kl_a}
L_{KL,a} = \mathbb{E}[KL(P(z_a')||\mathcal{N}(0,1))], z_a = E^{fc}_a(z_a'),
\end{equation}
where $E^{fc}_c$ denotes the final fully connected layer of content encoder $E_c$, and $E^{fc}_a$ denotes the final fully connected layer of attribute encoder $E_a$.

To ensure the encoded $z_c$ and $z_a$ describing content and attribute information, respectively, we apply a classifier $C$ pre-trained on the attribute of interest to guide the learning of $E_a$. Thus, this guided loss is calculated as:
\begin{equation}
\label{eq_cls_guide}
L_{guide} = |C(x) - C(G(\tilde{z}_c, z_a))|,
\end{equation}
where $C(\cdot)$ indicates the classifier. We note that, $\tilde{z}_c$ denotes a randomly sampled content feature, $\tilde{z}$ is sampled from $\mathcal{N}(0,1)$ and then is passed through the final fully connected layer of $E_c$. Thus, we have $\tilde{z}_c = E^{fc}_c(\tilde{z})$, and the image with identical attribute but random content can be produced as $G(\tilde{z}_c, z_a)$. 

From~\eqref{eq_cls_guide}, we see that the enforcement of classification output similarity between an input image $x$ and a synthesized one with the same $z_a$ yet with a random content $\tilde{z}_c$, would ensure our $E_c$ and $E_a$ to extract attribute-invariant and attribute-dependent representations, respectively. 
That is, the deployment of the classifier $C(\cdot)$ in Fig.~\ref{fig:archi} would guide the attribute encoder $E_a$ to extract attribute-dependent information by equation~\eqref{eq_cls_guide}. With equation~\eqref{eq_rec} ensuring the quality of reconstruction, attribute-invariant information would be encoded by content encoder $E_c$ for fair reconstruction.

\subsection{Enforcing Content and Attribute Consistency}

With the above guidance of the attribute classifier and the use of generative models, we have $E_a$ extract latent attribute features. With this classifier to be replaced by those pre-trained on preferable attributes of interests, one can easily extend the above architecture to disentangle the corresponding attributes. To further ensure our decomposed $z_c$ and $z_a$ from $z$ contain only content and attribute information, respectively, we advance feature consistency losses during the training of our dec-GAN. This is achieved by minimizing the content and attribute feature consistency loss defined as follows:
\begin{equation}
\label{eq_c_const}
L^{const}_{c} = |E_c(G(z_c, \tilde{z}_a)) - z_c|,
\end{equation}
\begin{equation}
\label{eq_a_const}
L^{const}_{a} = |E_a(G(\tilde{z}_c, z_a)) - z_a|.
\end{equation}
As illustrated in Figure~\ref{fig:archi}, $G(z_c, \tilde{z}_a)$ indicates the synthesized image with the same content as that of input $x$ but with different attributes $\tilde{z}_a = E^{fc}_a(\tilde{z})$. Similarly, we have $G(\tilde{z}_c, z_a)$ denote the generated image with the same attributes as those of $x$ but with different content information via $\tilde{z}_c$. By observing the above feature consistency, both $E_c$ and $E_a$ would extract associated content and attribute features, realizing the decomposition of $z$ into $z_c$ and $z_a$, respectively.

\section{Experiment}\label{exp}
We consider image datasets of MNIST~\cite{lecun1998gradient} and CMU Multi-PIE~\cite{gross2010multi} for our experiments. The former consists of 60,000/10,000 training/test digit images of 10 classes, while the latter contains face images with multiple viewpoint, illumination and expression variations. We only use a subset of CMU Multi-PIE with 5 viewpoints and smiling expression variation, which consists of 68,810 images.

For the generator to be decomposed, since our proposed architecture does not limit the use of particular GAN models, we first follow the backbone of VAE-GAN~\cite{larsen2015autoencoding}. In addition, we consider a second generative model with a deeper backbone~\cite{lee2018drit} and refer Res-GAN to this generative model. The encoder and the generator of Res-GAN consist of convolution layers and residual blocks. For the detail of the architecture of VAE-GAN and Res-GAN, please refer to the supplementary material.
For $E_c$ and $E_a$ in our dec-GAN, we simply utilize the same encoder structure of the model to be decomposed.
\subsection{Image Generation and Manipulation}

\subsubsection{MNIST}

\begin{figure}[t]
\centering
\includegraphics[width=\linewidth]{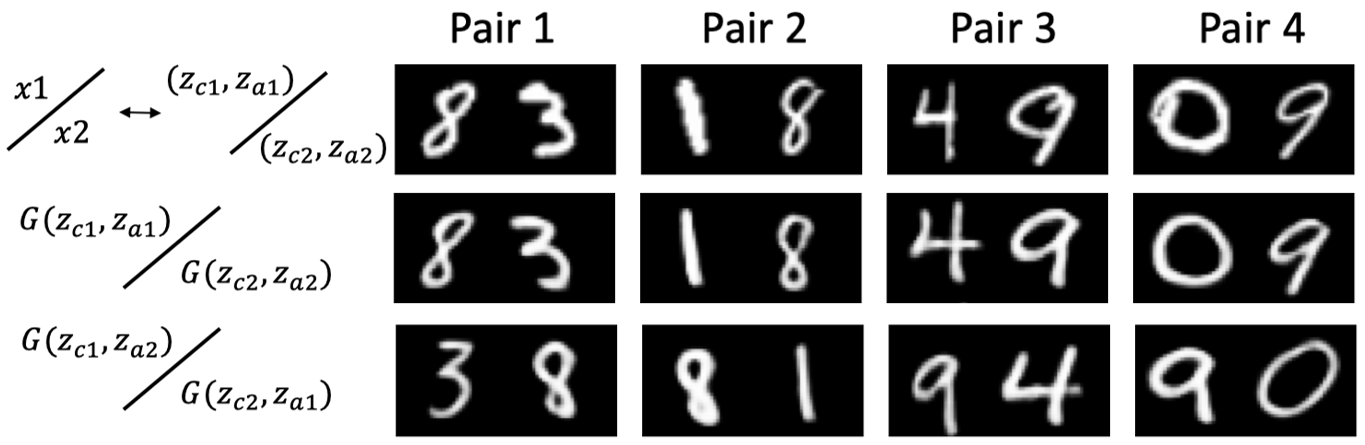}
\caption{Image generation via attribute swapping on MNIST. Note that $x$ indicates the input image, with the outputs $G$ are produced by the associated $z_c$ and $z_a$.} 
\label{mnist}
\end{figure}

For MNIST, the classifier $C$ is pre-trained to identify the digit categories, which are viewed as the attributes while the visual appearance like stroke thickness or angle as the content features. As shown in Figure~\ref{mnist}, we demonstrate our image generation results using different pairs of content and attribute features. The first row in Figure~\ref{mnist} show input image pairs, and the second row depicts reconstructed outputs using derived $z_c$ and $z_a$ features. Image outputs by swapping $z_c$ and $z_a$ are shown in the third row. From this row, we see that the synthesized image would preserve the same content as those in the first two rows, while the attribute (digit category) would match the other one in the input image pair. This confirms the effectiveness of our dec-GAN in disentangling content and attribute features, while the latter is guided by a digit classifier in this case. To further verify $z$, $z_c$ and $z_a$ capture different visual information, we conduct t-SNE visualization on such features using MNIST. Due to space limitation, such visualization results are presented in the supplementary material.

\subsubsection{CMU Multi-PIE}

\begin{figure*}[t]
\centering
\includegraphics[width=0.9\textwidth]{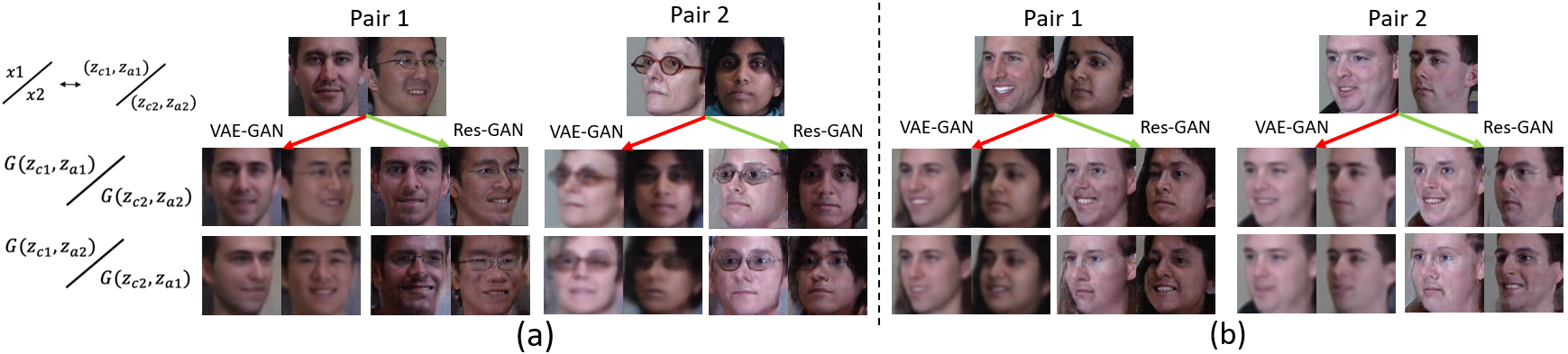}
\vspace{-0.3cm}
\caption{Image generation from CMU-MultiPIE via swapping the attributes of (a) pose and (b) smile. The first row shows sampled input image pairs $x_1$ and $x_2$, the second row shows reconstructed image outputs $G(z_c,z_a)$ of the input images, and the third row depicts generated image outputs by swapping $z_a$ in each pair. Comparing the second and the third row, we see that the image content is preserved while the attributes (i.e., pose/smile information) are swapped within each image pair. Note that results using VAE-GAN and Res-GAN as the backbones of our dec-GAN are shown.}
\label{multipie}
\end{figure*}

We take both VAE-GAN and Res-GAN as the backbone of our dec-GAN, and consider pose and expression (smile) as two distinct attributes of interest. We demonstrate image generation results when taking pose categories as attribute of interest in Figure~\ref{multipie}(a). The first row in Figure~\ref{multipie}(a) show input facial image pairs, and the second row depict reconstructed image outputs using derived $z_c$ and $z_a$. Image outputs by swapping $z_c$ and $z_a$ are shown in the third row. Comparing this row and the first two rows, we see that the manipulated facial images remained the same identity with pose information altered and matched the other one in the input image pair. 
Compared to discrete categorical attributes in MNIST, this confirms that our dec-GAN is able to handle continuous attributes such as poses.

In addition, we show the results when taking smiling expression as attribute of interest in Figure~\ref{multipie}(b). Similarly, by comparing the last row and the first two rows, we see that the manipulated images remained same facial information with only smiling expression altered. This confirms that smiling expression is able to be decomposed from the original latent feature. It is worth noting that we decompose pose and smiling attributes from the same pre-trained generative models, confirming the flexibility of our dec-GAN in extracting the attributes of interest.
\subsection{Quantitative Results}

\subsubsection{Quantitative Evaluation of $z_c$ and $z_a$}

We conduct quantitative experiments to examine the effectiveness of our dec-GAN in disentangling content and attribute features. With the use of CMU Multi-PIE face dataset, $z_a$ derived by our model would be expected to contain pose information only, while $z_c$ represents pose-invariant identity features. We then take these two types of features and perform pose and ID classification tasks, and compare the results to the uses of latent representations $z$ derived by VAE-GAN~\cite{larsen2015autoencoding} and UFDN~\cite{liu2018ufdn}.

Table~\ref{quant_table} lists and compares classification results of different tasks using $z$, $z_c$ and $z_a$. We simply apply a two-layer classifier (i.e., 2 fully-connected layers with ReLU activation, followed by a softmax layer) for comparison purposes. We do not apply additional or complex classifiers, which can possibly further improve the recognition performances. The number of classes is 5 for pose classification and 249 for identity classification. From this table, we observe that $z_a$ yielded the best result in pose classification, while $z_c$ resulted in the highest performances for identity classification. This is expected, since our dec-GAN is particularly designed to disentangle attribute-dependent and attribute-invariant features. Note that the use of $z$ of VAE-GAN achieved inferior results, indicating that its latent representation would contain both content and attribute information, and thus cannot be expected to sufficiently address either task. For UFDN, since they derive hand-crafted one-hot vector for attribute feature, their model is not applicable for pose classification.

\begin{table}
\centering
{\centering
    \begin{tabular}{c|c|c}
		\hline
		Method & Pose & ID \\
		\hline\hline
		VAE-GAN~\cite{larsen2015autoencoding} & 97.44 ($z$) & 96.94 ($z$)\\
		UFDN~\cite{liu2018ufdn} & N/A & 94.31 ($z_c$) \\
		Ours & \textbf{99.74} ($z_a$) & \textbf{98.59} ($z_c$)\\
		\hline
	\end{tabular}
}
\caption{Classification performances on CMU Multi-PIE. Note that our method decomposes content and attribute features ($z_c$ and $z_a$) from latent representation $z$ derived by pre-trained VAE-GAN. Since the attribute feature of UFDN~\cite{liu2018ufdn} is a hand-crafted one-hot vector, it cannot be directly applied for pose classification.}
\label{quant_table}
\end{table}
\vspace{-0.2cm}
\subsubsection{Comparisons of Training Time}

As noted earlier, a major advantage of our dec-GAN is the applicability to existing GAN-based models without the need of pre-defined attributes. We compare the numbers of training iterations and computation times of dec-GAN and ACGAN~\cite{odena2017icml} with the same backbone structures for generators and discriminators. Note that all experiments were conducted on single NVIDIA GTX 1080 Ti with batch size = 12 and the table of results are presented in the supplementary material.

We found that dec-GAN is four times faster than AC-GAN when disentangling smiling attributes (i.e., 6 vs. 26 mins) and is about seven times faster when disentangling pose attributes (i.e., 13 vs. 89 mins). This is because that the training of our dec-GAN can be initialized by existing GAN models, followed by the training of $E_c$ and $E_a$. On the other hand, AC-GAN needs to pre-define additional dimensions for describing the attribute, so that its training can not be initialized. From the above experiments, the flexibility and effectiveness of our dec-GAN can be confirmed.
\vspace{-0.1cm}
\section{Conclusion}
In this paper, we proposed a unique decomposition-GAN (dec-GAN) to perform feature disentanglement, which jointly extracts content and attribute representations from the latent feature observed from existing GAN-based models.
Different from prior disentanglement works which typically derive disjoint latent representations describing desirable features, our dec-GAN performs feature decomposition, which separates latent representation into separate features describing the properties/attributes of interest. The attribute disentanglement of our dec-GAN is driven by classifiers pre-trained on the attribute of interest. Followed by the design of generative network modules, this allows disentanglement of content and attributes, while exhibiting additional flexibility in determining the attributes of interest (i.e., by replacing such classifiers based on the desirable attribute categories). We performed qualitative and quantitative evaluation using multiple image datasets, with attributes ranging from digit categories to pose angles. The effectiveness and robustness of our dec-GAN can be successfully confirmed, while its superiority over existing models can also be verified.\\

\noindent\textbf{Acknowledgement}
This work is supported in part by the Ministry of Science and Technology of Taiwan under grant MOST 110-2634-F-002-036. We also thank to National Center for High-performance Computing (NCHC) for providing computational and storage resources.
\bibliographystyle{IEEEbib}
\bibliography{strings,refs}

\end{document}